\documentclass[sigconf]{acmart}

\def\BibTeX{{\rm B\kern-.05em{\sc i\kern-.025em b}\kern-.08emT\kern-.1667em\lower.7ex\hbox{E}\kern-.125emX}}

\copyrightyear{2019} 
\acmYear{2019} 
\setcopyright{acmlicensed}
\acmConference[KDD '19]{The 25th ACM SIGKDD Conference on Knowledge Discovery and Data Mining}{August 4--8, 2019}{Anchorage, AK, USA}
\acmBooktitle{The 25th ACM SIGKDD Conference on Knowledge Discovery and Data Mining (KDD '19), August 4--8, 2019, Anchorage, AK, USA}
\acmPrice{15.00}
\acmDOI{10.1145/3292500.3330756}
\acmISBN{978-1-4503-6201-6/19/08}

\usepackage{multirow}
\usepackage{subcaption}
\usepackage{listings}
\usepackage{url}
\usepackage{todonotes}
\usepackage{threeparttable}

\begin{document}

%
\title{Chainer: A Deep Learning Framework for  Accelerating the Research Cycle}

\author{Seiya Tokui, Ryosuke Okuta, Takuya Akiba, Yusuke Niitani, Toru Ogawa, Shunta Saito,}
\author{Shuji Suzuki, Kota Uenishi, Brian Vogel, Hiroyuki Yamazaki Vincent}
\email{tokui@preferred.jp}
\affiliation{%
  \institution{Preferred Networks, Inc.}
  \streetaddress{3F Otemachi Bldg., 1-6-1 Otemachi, Chiyoda-ku}
  \city{Tokyo}
  \country{Japan}
  \postcode{100-0004}
}

%
\renewcommand{\shortauthors}{Tokui, et al.}

%
\begin{abstract}
Software frameworks for neural networks play a key role in the development and
application of deep learning methods. In this paper, we introduce the Chainer framework, which intends to provide a flexible, intuitive, and high performance means of implementing the full range of deep learning models needed by researchers and practitioners. Chainer provides acceleration using Graphics Processing Units with a familiar NumPy-like API through CuPy, supports general and dynamic models in Python through Define-by-Run, and also provides add-on packages for state-of-the-art computer vision models as well as distributed training.
\end{abstract}

%
%


%

%

%
\maketitle

\section{Introduction}


Deep learning is driving the third wave of artificial intelligence research~\cite{mafia}.
Recent investigations indicate that deep learning is moving beyond its early successes in pattern recognition and toward new applications in diverse domains and industries.
To implement these research ideas, a software framework for deep learning is required.

Implementing neural networks (NNs) requires a set of specialized building blocks, including multidimensional arrays, activation functions, and automatic differentiation.
To avoid duplicating these tools, many developers used open-source deep learning frameworks such as Caffe~\cite{Jia13caffe} or Torch~\cite{collobert:2008a}.
Because deep learning was first used successfully in computer vision and speech recognition, early deep learning frameworks were designed primarily for feed-forward networks such as convolutional neural networks (CNNs), which are effective for analyzing fixed-length data such as images.

More recently, additional types of deep learning models have become a major research topic.
Following the impressive results in game playing~\cite{mnih2013playing}, deep reinforcement learning has become a promising research area.
In addition, after recurrent neural networks (RNNs) demonstrated promising results on variable-length data such as text, the use of these models has increased.
RNNs with Long Short-Term Memory (LSTM) are currently being used with success for machine translation~\cite{conf/nips/SutskeverVL14} and conversation models~\cite{journals/corr/VinyalsL15}. 

However, as most of the existing deep learning frameworks are designed for image processing using CNNs, they lack support for abstracting data structures and training models to implement more general deep learning models.
In addition, many existing frameworks use a domain-specific language for representing deep learning models, along with an interpreter to translate them into a data structure stored in memory.
Therefore, developers using these frameworks cannot use standard programming language debuggers--a significant problem as debugging is a major aspect in developing and tuning deep learning models.     

We herein introduce Chainer, an open-source framework for deep learning that provides a simple and efficient support for implementing complex algorithms, training models, and tuning model parameters.
The remainder of the paper is organized as follows.
Section 2 describes the standard architecture of the existing deep learning frameworks.
Section 3 introduces the architecture of Chainer.
Section 4 describes the performance techniques such as memory usage optimizations and double backpropagation techniques.
Section 5 presents CuPy as a backend library for Graphics Processing Units (GPUs).
Section 6 describes distributed training capability.
Section 7 introduces ChainerCV, an add-on package for computer vision.
Section 8 presents the related work.
Finally, Section 9 provides a summary and the directions for future work.

\section{Background and Motivation}


\begin{figure*}
\centering
\begin{subfigure}{.5\linewidth}
  \centering
  \includegraphics[width=.75\textwidth,clip]{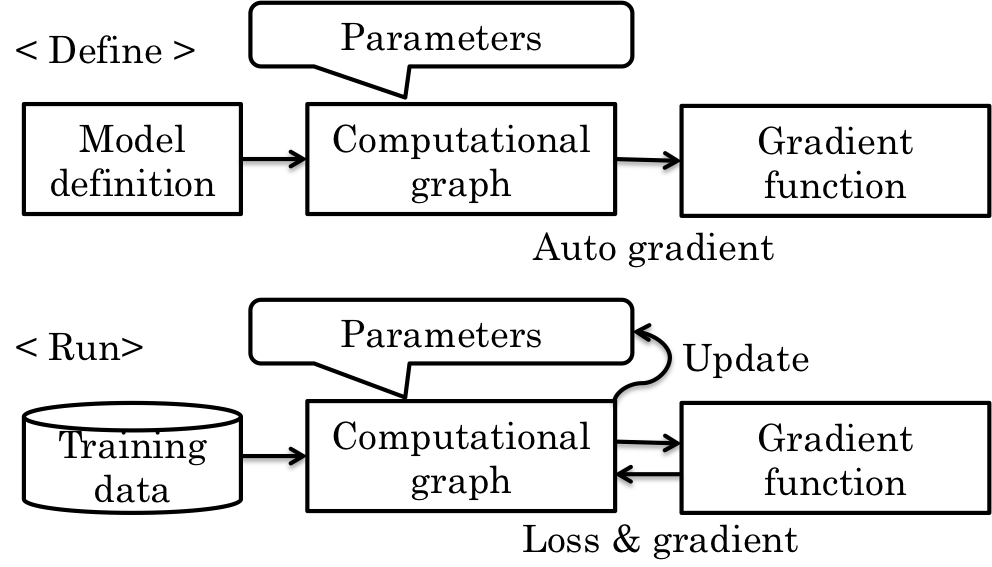}
  \caption{\textit{Define-and-Run}: existing approach}
  \label{fig:define-and-run}
\end{subfigure}%
\begin{subfigure}{.5\textwidth}
  \centering
  \includegraphics[width=.75\textwidth,clip]{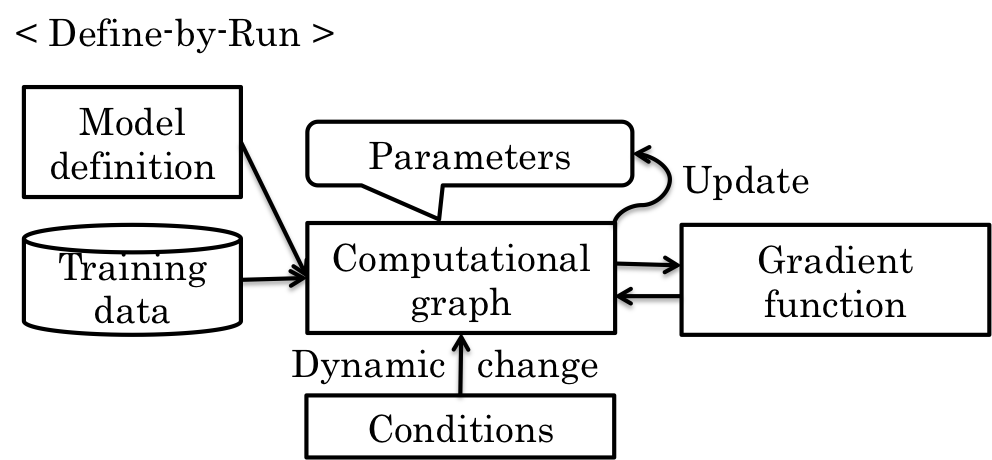}
  \caption{\textit{Define-by-Run}: new approach}
  \label{fig:define-by-run}
\end{subfigure}
  \caption{Relationship between computational graph construction and training.}
  \label{fig:paradigm}
\end{figure*}

In typical NN frameworks, models are built in two phases, in a paradigm that we name as \textit{Define-and-Run} (Figure~\ref{fig:define-and-run}).
In the Define phase, the computational graph of the model is first defined and constructed.
This phase corresponds to the instantiation of a neural network object based on a model definition that specifies the data flow graph of inter-layer connections, initial weights, and activation functions.
Automatic differentiation is typically used to define the computations for both the forward and backward passes, with optional graph optimizations being performed as well.
In the Run phase, the actual forward and backward calculation of the graph is performed.
Provided a set of training examples, the model is trained in this phase by minimizing the loss function using optimization algorithms such as stochastic gradient descent.

Under the Define-and-Run paradigm, static NN models such as CNNs can be implemented easily.
The model definition may be written in a specific markup language such as Protobuf or YAML~\cite{journals/corr/GoodfellowWLDMPBBB13}.
The deep learning framework serves as an interpreter that executes the model definition, which can be regarded as an independent NN program.
The NN program receives the inputs (data examples), processes them (forward/backward computation), changes the model's internal state (updating), and outputs the results (predictions).

The Define-and-Run paradigm operates well for static models such as CNNs because having the full computational graph available enables potential graph optimizations to improve the memory efficiency and/or runtime performance.
However, for implementing other types of NN models, two major problems arise.

The first is that it can be cumbersome to support general dynamic graphs, i.e., neural networks with control flow.
In frameworks such as TensorFlow~\cite{tensorflow2015-whitepaper}, the control flow decisions are defined in the data flow graph using special operators such as \textit{Switch} and \textit{Merge} rather than using the control flow syntaxes of the host language.

The second problem is that under the Define-and-Run paradigm, the inner mechanism of the neural network is not accessible to the user.
This presents several difficulties in the creation of an effective model.
For example, to debug and tune a model effectively, a user must be able to observe what is occurring inside the model.
However, as a large object of a single class, the computational graph contains the entire model's information, i.e., its structure, weights, gradients, and internode operations, implying that it is essentially a black box.
Consequently, development tools such as profilers and debuggers cannot determine the model's faults or how it could be improved.
If graph optimizations are performed by the framework, this problem is compounded further.

\section{Design and Programming Model}

In this section, we present the basic design of automatic differentiation APIs based on the Define-by-Run paradigm (Figure.~\ref{fig:define-by-run}).

\subsection{On Demand Graph Construction}

Backpropagation is executed by bookkeeping the history of operations applied to the input arrays and backtracking the history.
In the Define-by-Run paradigm, the history of operations is recorded simultaneously with the forward computation applied to concrete input arrays.
This can be achieved by creating a node in the computational graph for each variable and operation.
The computational graph only defines how to backtrack the operations applied to the input, and does not define the forward computation.
We define the computational graph by two types of nodes: \emph{variable nodes} that represent the variables involved in the computation, and \emph{function nodes} that represent the operations applied to the variables.
After applying a function $f$ to the input variable $x$ and obtaining an output variable $y$, the function node $n_f$ contains a reference to the input node $n_x$, and the output node $n_y$ contains a reference to the function node $n_f$.
These references are used to backtrack the graph.

The program that defines the forward computation is similar to any standard numerical computations that do not compute any gradients.
The only difference is that each differentiable function stores its computational history into the graph in addition to computing its output.
Because the graph construction only relies on the execution trace of the program, it can be combined with arbitrary syntactic constructs of the host language, e.g., conditional branches and loops.
Such a program generates a graph with a different topology and size at each invocation, while maintaining the correct gradient computation.
The power of the host language that we can use is not limited to such primitive language constructs; we can also leverage high-level tools such as debuggers and profilers.

\begin{figure}
  \begin{center}
    \begin{subfigure}{\linewidth}
    \begin{verbatim}class Linear(Link):
    def __init__(self, n_in, n_out):
        with self.init_scope():
            self.W = Parameter(HeNormal(),
                               (n_out, n_in))
            self.b = Parameter(0, (n_out,))

    def forward(self, x):
        return x @ self.W.T + self.b

class MultiLayerPerceptron(Chain):
    def __init__(self, n_in, n_hid, n_out):
        with self.init_scope():
            self.l1 = Linear(n_in, n_hid)
            self.l2 = Linear(n_hid, n_out)

    def forward(self, x):
        h = relu(self.l1(x))
        return self.l2(h)\end{verbatim}
    \end{subfigure}    
  \end{center}
  \caption{Examples of model definitions by object-oriented programming.}
  \label{fig:link-chain}
\end{figure}

\subsection{Object-Oriented Model Definition}

Compositionality is an important characteristic of deep learning.
Fragments of networks are connected in various combinations to form a rich set of architecture.
APIs to write deep models should exhibit compositionality to reuse and combine components flexibly.

In the Define-by-Run paradigm, models consist of the code defining the forward computation and parameters deciding its behavior.
The code is written as a host language program, and must be bound to parameters.

This parameter-binding problem is resolved by object-oriented programming.
Each neural network fragment that involves its own parameters is defined by a class.
Such fragments are combined into another class to create larger model components.
Hence, the modularization of neural networks and parameter binding are resolved.
Figure~\ref{fig:link-chain} shows an example of defining a fully connected layer and a multilayer perceptron.
The parameters are initialized at object construction, and the forward computation is written as a method.

Object-oriented model definition also provides a unified interface to models in terms of parameter handling.
Because the model is composed of a tree of model fragments, the parameters of specific subtrees can be collected easily by traversing it.

This style of object-oriented model definition was first introduced by Chainer in 2015, and is now widely used in other Define-by-Run frameworks, e.g., PyTorch and TensorFlow Eager.

\section{Technical Features}

In this section, we describe several techniques in Chainer that improve its simplicity and efficiency, applicable to frameworks based on the Define-by-Run paradigm.

\subsection{Memory-Efficient Backpropagation} \label{tech:memory}

Memory efficiency is of central interest in deep learning frameworks as the sizes of models and data are limited by the amount of available physical memory.
Optimization can be performed further at the framework level, especially in reducing peak memory usage.

\subsubsection{Global Memory Usage Reduction}

In deep learning frameworks based on the Define-by-Run paradigm, memory management is naturally delegated to that of the host language.
Chainer relies on reference-counting garbage collection (GC), which is the primary mechanism for memory management in the standard Python implementation (a.k.a. CPython).
Automatic differentiation APIs based on the Define-by-Run paradigm operates well with reference-counting GC; a subgraph of the computational history is released immediately once it is rendered unreachable.

When a graph is released by reference-counting GC, each node is released in the topological order of the computational graph.
Meanwhile, the backpropagation algorithm visits the nodes in the topological order.
By merging these two procedures, we can minimize the peak memory consumption of backpropagation.
This is accomplished by manually eliminating the reference to a function node immediately after processing it during graph backtracking.

\subsubsection{Function-wise Local Memory Usage Reduction}

In general, the gradient (or more precisely, the Jacobian matrix) of a function depends on the input.
Therefore, it is natural to design the interface of the backward computation such that it takes both the input arrays and output error as arguments.
This interface, however, prevents us from applying memory usage optimization for operations that do not require the input arrays to compute the gradient.
Some operations, e.g., $\tanh$, can use the output arrays instead of the input arrays to compute the gradient.
When the next operation applied to the output requires the inputs for gradient computation, we can eliminate the input arrays and retain the output arrays such that the data kept on memory for backpropagation are minimized.
Further, some operations require neither the input nor the output arrays.
In this case, we can eliminate the references to both of them.

Each differentiable operation is implemented as a subclass of \texttt{FunctionNode} with overridden \texttt{forward} and \texttt{backward} methods.
In \texttt{forward}, the inputs and outputs required for \texttt{backward} are explicitly declared through the \texttt{retain\_inputs} and \texttt{retain\_outputs} method calls.
If an input or output is not listed by these declarations, that input/output is not saved for backpropagation.
In particular, inputs are no longer passed to the \texttt{backward} method; instead, the implementation of \texttt{backward} pulls them only when necessary.

Chainer utilizes a particular variable object representation that is designed to release memory as soon as possible once it is no longer required.
In a naive implementation of automatic differentiation with the Define-by-Run paradigm, each variable node would directly contain a multidimensional array (for example, as an attribute).
An issue with such a design is that the memory used by the array cannot be reclaimed until the last reference to the variable has been deleted.
In particular, even if the user code does not hold any direct references to the variable, the computational graph may still hold a reference to it, in which case its memory cannot be reclaimed.
This issue arises owing to the inability of distinguishing user code references from internode references.
It is noteworthy that provided user code references to a variable are alive, it is necessary to maintain the multidimensional array data associated with the variable.
Meanwhile, the references inside the computational graph do not always require the data to be alive; if no operation retain the variable as an input or output, the data should be released.
Based on this observation, we can resolve this issue using separate objects to represent the variables in the user code and the variables in the computational graph.
In Chainer, each \texttt{Variable} object holds the array data, and is distinct from the corresponding \texttt{VariableNode} object representing the variable node in the graph.
The variable node object holds a reference to the array data only when the variable is retained by an operation.
With this formulation, we can immediately reclaim the memory for the variable once the last reference from the user code has been removed, unless an operation retains it as an input or output.

\subsection{Double Backpropagation}

Backpropagating through computation involving gradient computation is a major feature of modern deep learning frameworks.
It corresponds to automatic differentiation for Hessian-vector product.
Such a feature is sometimes called \textit{double backpropagation}.

Double backpropagation is supported by implementing the backward computation of each operation using functions supporting differentiation.
Although this idea may appear straightforward, a naive implementation may result in reference cycles that cause unnecessary memory consumption.
Two factors must be considered to avoid reference cycles: interface to access the resulting gradients, and output retention at each differentiable function.

\subsubsection{Interface to Access the Resulting Gradients}

Two styles of interface exist to trigger backpropagation.
The first one is the \break \texttt{Variable.backward()} method, which computes the gradient with respect to each input.
The resulting gradients are stored directly in the variable objects.
The other one is the \texttt{grad()} function that takes both a set of inputs and a set of outputs as arguments.
In this case, the function returns the set of computed gradients corresponding to the specified outputs.
The latter interface does not introduce additional references between objects, while the former may add references from the input nodes to the computed gradients.
Because the computed gradients refer the input nodes indirectly through the computational graph, a reference cycle appears.

This reference cycle is removed by discriminating between user code references and inter-node references, as discussed in Section \ref{tech:memory}.
Because the reference from a variable to the corresponding gradient is not part of the computational graph, we place this reference into the \texttt{Variable} object instead of into the \texttt{VariableNode} object.

\subsubsection{Output Retention for Double Backpropagation}

As detailed in Section \ref{tech:memory}, the \texttt{backward} method of each \texttt{FunctionNode} implementation may use the output variables declared to be retained in the forward computation.
To render backpropagation differentiable, a special step is required because the function node cannot maintain a reference to the output node; otherwise, a reference cycle is introduced.
It entails that the output node may be released before the backward computation of the function node is executed.
We can still maintain the validity of the differentiable backpropagation by replaying the graph construction for such an output node, i.e., a fresh node object is created and connected during backpropagation as if it were the output node.
Further, we store the output array data to the function node as a backup, and use them for the recreated output node.
This does not nullify the computational validity; the output node being released indicates that no other nodes or user codes contain any references to it; therefore, recreating the output node does not conflict with any existing nodes.

\section{GPU Support by CuPy}

The typical usage of a deep neural network requires significant power for floating point numeric calculation; therefore, it is necessary for deep learning frameworks to fully leverage the computing power of external accelerator such as GPUs.
This is not trivial for people who write deep neural network codes to implement high performance GPU programs while maintaining its flexibility, simplicity, and ease in extending components.
CuPy is an open-source library for Python that provides the computational power of NVIDIA GPUs with the NumPy-compatible syntax.
It accelerates any computation described in a NumPy-like syntax by fully utilizing the GPU architecture with the CUDA platform provided by NVIDIA, including cuBLAS, cuDNN, cuRAND, cuSOLVER, cuSPARSE, and NCCL.

The interface of CuPy is highly compatible with that of NumPy; in most cases, it can be used as a drop-in replacement.
It supports standard numerical data types, array indexing, slice, transpose, reshape, and broadcasting.

Users can create custom CUDA kernels to execute codes faster, using code snippets of C++.
CuPy automatically wraps and compiles the code to create a CUDA binary.
Compiled binaries are cached and reused in subsequent runs.

CuPy was first developed as the backend of Chainer.
The initial version of Chainer was implemented using PyCUDA\cite{pycuda}, a widely used Python library for CUDA GPU calculation.
However, PyCUDA could not support enough functionalities of NumPy for deep learning and the CUDA support was insufficient.
CuPy became independent from Chainer in June 2017, when Chainer v2.0 and CuPy v1.0 were released.
Henceforth, numerous non-deep-learning projects have leveraged CuPy's strong performance and simple interface.
For example, a Python-based probabilistic modeling software, Pomegranate\cite{pomegranate2017}, and a natural language processing library, spaCy\cite{spaCy}, use CuPy as their GPU backend.

\subsection{CuPy Example}

Because CuPy is a Python package similar to NumPy, it can be imported into a Python program similarly.
As shown in Fig.~\ref{fig:norm_code} code, \texttt{cp} is used as an abbreviation of CuPy, similar to \texttt{np} for NumPy.
The \texttt{cupy.ndarray} class is in the core of CuPy as a GPU alternative of \texttt{numpy.ndarray}.
In the code, \texttt{x} is an instance of \texttt{cupy.ndarray}. Its creation is identical to the NumPy syntax, except that NumPy is replaced with CuPy.
The primary difference of \texttt{cupy.ndarray} from \texttt{numpy.ndarray} is that the content is allocated on the GPU memory.
Most of the CuPy array manipulations are similar to those of NumPy.
For example, NumPy uses \texttt{numpy.linalg.norm} to calculate the Euclidean norm on a CPU, while CuPy uses \texttt{cupy.linalg.norm} to calculate it on a GPU.

\begin{figure}
  \begin{center}
    \begin{subfigure}{.22\textwidth}
       \begin{verbatim}import numpy as np
x = np.array([1, 2])
l2 = np.linalg.norm(x)\end{verbatim}
      \caption{NumPy}
      \label{fig:norm_numpy}
    \end{subfigure}  
    \begin{subfigure}{.22\textwidth}
      \begin{verbatim}import cupy as cp
x = cp.array([1, 2])
l2 = cp.linalg.norm(x)\end{verbatim}
      \caption{CuPy}
      \label{fig:norm_cupy}
    \end{subfigure}
  \end{center}
  \caption{Examples using NumPy and CuPy.}
  \label{fig:norm_code}
\end{figure}

\subsection{Supported Functionalities}

As demonstrated in the previous subsection, CuPy implements many functions on \texttt{cupy.ndarray} objects.
See the reference~\footnote{https://docs-cupy.chainer.org} for the supported subset of NumPy API. 

\subsubsection{Linear Algebra}
CuPy supports most linear algebra functions in NumPy such as eigen decomposition, Cholesky decomposition, QR decomposition, singular value decomposition, linear equation solver, inverse of matrix, and the Moore-Penrose pseudo inverse.
These functions are defined in \texttt{cupy.linalg} and are compatible with \texttt{numpy.linalg}. All of them are backed by cuSOLVER, a LAPACK implementation that operates on GPUs.

\subsubsection{Sparse Matrices}
CuPy supports sparse matrices using cuSPARSE.
These matrices contain the same interfaces of sparse matrices in SciPy~\cite{jones2001scipy}, \texttt{scipy.sparse}.
Depending on their requirements, users can choose between coordinate-format sparse matrix, compressed sparse row matrix, compressed sparse column matrix, or sparse matrix with diagonal storage.

\subsubsection{Sorting}
CuPy provides sort, argsort, and lexsort functions that are compatible with NumPy, backed by Thrust, a library of parallel algorithms written in C++ with CUDA.
CuPy takes the advantage of Thrust, which implements sophisticated parallel sort algorithms for GPUs.

\begin{figure}
  \begin{center}
    \begin{verbatim}kernel = cupy.ElementwiseKernel(
    'float32 x, float32 y, float32 z',  # arguments
    'float32 w',                        # outputs
    'w = x * y + z',                    # computation
    'my_mad')                           # kernel name
w = kernel(x, y, z)\end{verbatim}
  \end{center}
  \caption{Example of a user-defined kernel.}
  \label{fig:cupyudf}
\end{figure}

\subsection{Custom CUDA Kernels}

CuPy is easy to extend with user-defined kernels by combining operators, two types of kernels, and generic types.
It is easy to compose and launch an arbitrary kernel in GPUs with the CUDA code fragments.

The element-wise kernel applies the same operation to all elements.
For example, the \texttt{cupy.add} function applies the $+$ operator for each element pair.
The reduction kernel folds all elements by a binary operator.
For example, the \texttt{sum} function folds all elements by the $+$ operator.
Element-wise kernels and reduction kernels are analogous to Map and Reduce from MapReduce~\cite{dean2004}, respectively.
Figure \ref{fig:cupyudf} is an example of a user-defined element-wise kernel.
The first and second arguments comprise a list of input variables and a list of output variables, respectively.
The definition of each variable consists of the type specifier and the name of an argument.
The third argument is a CUDA code snippet that the user wants to define.
In the code snippet, an arbitrary CUDA code can be used.

CuPy also supports generic types.
With type parameters such as \texttt{T} specified instead of concrete types such as \texttt{float32}, custom kernels are generated as template functions.
Arguments with generic types accept arbitrary types of arrays.
For example, when the input type is specified as \texttt{'T x, T y'}, this function takes a pair of arrays with the same arbitrary data type such as integer and float.

\section{Distributed Parallel Training}

In this section, we introduce the distributed learning capability
component of Chainer, formerly called \emph{ChainerMN}.

Although the GPU performance has improved continuously, the training
process is still time consuming even with latest GPUs.
For example, training ResNet-50~\cite{He2016} for the ImageNet
dataset~\cite{imagenet2009} typically takes as long as one week with
a single GPU.
Chainer's distributed capability allows integrating power of
multiple GPUs fully utilizing hardware performance while preserving
Chainer's flexibility enabled by its Define-by-Run approach.
This allows for easy distributed learning even in complex use cases
such as dynamic neural networks, generative adversarial networks,
and deep reinforcement learning.

\subsection{Basics of Distributed Deep Learning}
\label{sec:chainermn:preliminaries}

\subsubsection{Data and Model Parallelism}

Two primary approaches are available to parallelize training by
distributed processing: data parallelism and model parallelism.
In data parallelism, each worker has a model replica and calculates
the gradients of different minibatches.
Workers update their model with these gradients collaboratively.
If we
define the batch size processed by each worker as $b$ and the number
of workers as $n$, the gradient obtained through communication is
equivalent to that in the batch size $bn$.
With more workers gradients
are calculated with more training data in one iteration, thus the gradient quality is improved and accelerating the learning process.

In model parallelism, each worker has a portion of the model and
cooperates with others to calculate one minibatch ~\cite{DistBelief}.
Model parallelism had been actively adopted particularly when GPU memory was small.
Currently, data parallelism is shown to be more efficient,
but in case where a model has a huge number of parameters such as domain of
natural language processing, model parallelism is adopted in combination
with data parallelism~\cite{shazeer2018mesh}.

\subsubsection{Synchronous vs.~Asynchronous} \label{syncvsasync}

Design choice on communication model from two options, synchronous or asynchronous model is the key factor to construct the overall parallel computation.
Both models are described below, focusing on data parallelism.

Synchronous data parallelism in distributed training has one additional step called all-reduce step compared to non-parallelized training sequence that consists of forward computation, backward computation, and optimization.
All-reduce is a parallel computing operation where the sum of parameters is calculated and distributed over all processes.
This is a standard functionality of MPI.
In the additional all-reduce communication step, workers communicate with
each other to obtain and distribute the sum of gradients calculated by individual
workers. Each worker calculates the average of gradients by dividing the sum by the number of replicas, and updates its own replica of the model with the gradient obtained through the all-reduce communication before optimization.

The asynchronous model, meanwhile, has special workers called
parameter servers. The parameter server owns and controls the model
parameters during the training process. Normal workers send gradients to the parameter server once
the gradients are obtained by forward and backward calculations. The
parameter server receives and uses the gradients to update the
model. Workers receive new model parameters and start the calculation of the new gradients.

\subsection{Parallelism Design}
Chainer adopts data parallelism and the synchronous communication model.
In the following, we will explain the choice from the options discussed in Section~\ref{sec:chainermn:preliminaries}.

Data parallelism requires no changes but a few additions to existing implementation; splitting dataset and computing the gradient average among workers.
This is because data
parallelization is tantamount to increasing a minibatch size in many cases such as image recognition.
Thus, we first chose the data parallelism but later added experimental
support on model parallelism.
Further, We adopted the synchronous communication model for its deterministic behaviour and convergence~\cite{Xinghao2017,Goyal2017}.

Synchronous data-parallel gradient exchange can be realized by all-reduce
communication between workers and thus Chainer is potentially capable of
running on any communication library that supports the all-reduce operation
with Chainer's communicator abstraction.
The first library supported by Chainer is OpenMPI, which is especially efficient with a CUDA-aware build.
However, all-reduce communication especially requires efficiency
because it is called in every training iteration and needs to process
a large amount of data. We adopted NCCL~\cite{nccl} developed by NVIDIA
as a primary library to run all-reduce. NCCL is a
highly-optimized communication library which enables efficient
all-reduce operation between NVIDIA GPUs within and across
nodes.

\subsection{API Design}
We describe the design goal of a distributed learning capabilty of Chainer, followed by a
description of the minimal steps to extend an existing deep learning
program written in Chainer to support distributed training. 

The flexibility of Define-by-Run design of Chainer should not be sacrificed
for distributed execution. The Define-by-Run allows the model structure to differ between
iterations, only assuming that the model structures are
identical between workers merely in a single iteration for all-reduce. Communication
for gradient exchange occurs immediately before the optimization step,
which is transparent to other Chainer components. Within
the bound of this minimal assumption,
any code can be put before or after the optimization step and model structure
can be changed at any iteration dynamically.

Figure~\ref{fig:chainermn} shows a core part of a
program to train the MNIST classification model,
including three primary additions in distributed mode: 
\textit{(1)} a communicator component that controls all inter-process communication,
\textit{(2)} transforming optimizer to \texttt{mutli\_node\_optimizer} to exchange gradients among workers, and
\textit{(3)} scatter the dataset to all workers.

\texttt{mutli\_node\_optimizer} is the most important component in
making Chainer distributed. It wraps the normal optimizer and exchanges the
gradient across processes using the all-reduce operation before
optimizing the model. It behaves
identically as the original optimizer except for the communication.
At the final step, \texttt{scatter\_dataset} lets workers make consensus on which fragment of training data to read for data parallelism.
Training dataset are split into equal fragments and distributed over worker processes.

Requiring as minimal changes in porting to distributed mode as possible has
allowed Chainer to preserve its flexibility afforded by the Define-by-Run paradigm.

\begin{figure}
  \begin{center}
    \begin{subfigure}{\linewidth}
      \begin{verbatim}# (1) Create a communicator
comm = chainermn.create_communicator()

# (2) Create and use multi_node_optimizer
optimizer = chainermn.create_multi_node_optimizer(
    chainer.optimizers.Adam(), comm).setup(model)
    
# (3) Distribute a dataset
train = chainermn.scatter_dataset(
    train, comm, shuffle=True)

# Use Chainer's Trainer class to simplify
# a forward-backward-optimization loop
iterator = chainer.iterators.SerialIterator(
    train, args.batchsize)
updater = training.StandardUpdater(
    train_iter, optimizer, device=device)
trainer = training.Trainer(updater, (100, 'epoch'))\end{verbatim}
    \end{subfigure}    
  \end{center}
  \caption{Example of training code using ChainerMN.}
  \label{fig:chainermn}
\end{figure}

\label{sec:evalmn}
\subsection{Evaluation}

We used the 90-epoch ResNet-50~\cite{He2016} training on the ImageNet dataset as our benchmark.
This task has been extensively used in evaluating the performance of distributed deep learning~\cite{Goyal2017, You2017, Codreanu2017}.

We used a cluster of 128 nodes, each of which is equipped with eight NVIDIA Tesla P100 GPUs.
The per-worker minibatch size was 32 and the total minibatch size was 32k with 1024 workers.
Further details of the experimental setups are provided in the appendix.


%


Figure~\ref{fig:scaling} illustrates
the communication time (i.e., all-reduce operations)
and time to complete a whole iteration (i.e., forward and backward computation, communication, and optimization)
for different numbers of GPUs, averaged over 100 iterations.
Our scaling efficiency when using 1024 GPUs
was 70\% and 80\% in comparison to single-GPU and single-node (i.e., 8 GPUs) baselines, respectively.
Using 1024 GPUs,
the mean training time over five independent runs was $897.9 \pm 3.3$s for 90 epochs, including the validation after each epoch.

\begin{figure}
  \centering
  \includegraphics[width=0.8 \hsize]{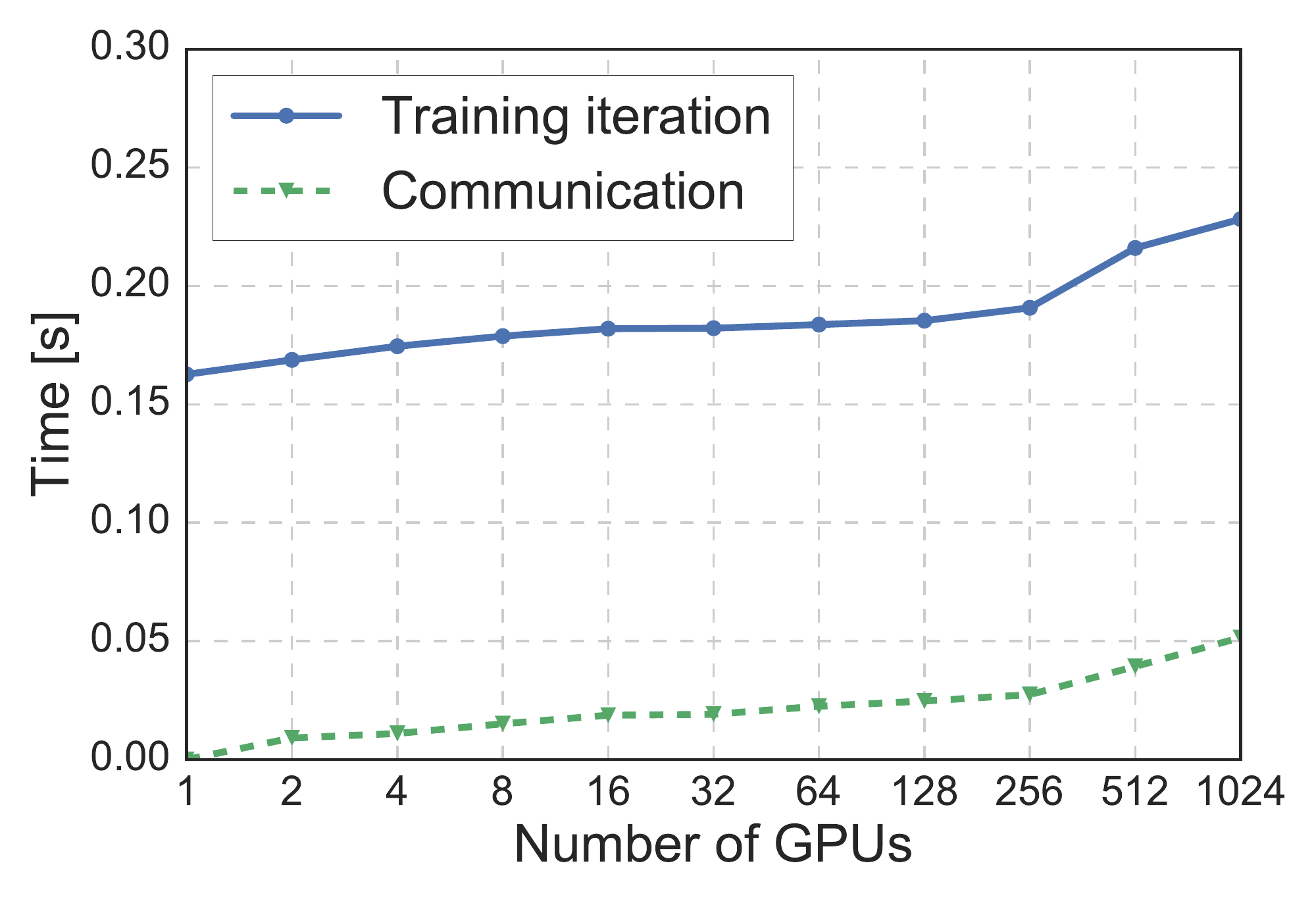}
  \caption{Iteration and communication time of ResNet-50 training for different numbers of GPUs.}
  \label{fig:scaling}
\end{figure}


\section{ChainerCV}
In this section, we introduce \textit{ChainerCV}, an add-on package for computer vision tasks.

Despite the powerful capability of Chainer, a gap still exists between what Chainer supports and what deep learning in computer vision requires.
Deep learning models have become increasingly stronger and more complex.
Therefore, it would be difficult for researchers and engineers to implement these algorithms from scratch.
Additionally, many typical computer vision utilities exist, such as data loaders and pre/post-processing functions such as non-maximum suppression~\cite{nms} that are outside the scope of Chainer.

ChainerCV aims at facilitating non-experts in the fast prototyping of ideas and the reduction in the barrier to enter the field.
The library provides state-of-the-art models, their pre-trained weights, and training scripts for various computer vision tasks such as image classification, object detection, semantic segmentation, and instance segmentation.
Additionally, the library provides utilities such as data loaders and evaluation metrics with a unified API.
Our design is based on the following three principles: \textit{easy-to-use}, \textit{unified API}, and \textit{reproducibility}.

\subsection{Easy to use}
ChainerCV supports four tasks: image classification~\cite{imagenet}, object detection~\cite{lin2014microsoft}, semantic segmentation~\cite{cityscapes}, and instance segmentation~\cite{fcis2017li}.
For each task, several neural networks may be implemented.
For instance, seven implementations are included for object detection.

Performing an inference with a ChainerCV's implementation is easy owing to a simple interface shared among models prepared for the same task.
Even for models solving the same task, they differ by the output type of the neural networks and how the outputs are post-processed.
The inference interface hides the difference in the underlying implementations among different models.
Additionally, the inference process is simplified further by automatically downloading pre-trained weights when a model object is instantiated.
Using pre-trained weights, an inference can be implemented in only two lines of the Python code, as shown in Figure~\ref{fig:chainercv_model}.

In addition to supporting an easy-to-use inference, ChainerCV supports training scripts that are easily customizable for a subset of models.
The training scripts are written using Chainer's training abstraction; therefore, training components can be swapped easily.
The scripts are designed to be extended by users, for instance, to train with a custom user dataset.
\begin{figure}[t]
  \begin{center}
    \begin{subfigure}{\linewidth}
      \begin{verbatim}# Instantiate object detection models
model_1 = FasterRCNNVGG16(pretrained_model='voc0712')
model_2 = SSD300(pretrained_model='voc0712')
  
# Make predictions
bboxes, labels, scores = model_1.predict([img])
bboxes, labels, scores = model_2.predict([img])\end{verbatim}
    \end{subfigure}    
  \end{center}
  \caption{Example of executing inference with two object detection models. The two models contain the same interface to conduct an inference.}
  \label{fig:chainercv_model}
\end{figure}

\subsection{Unified API}
ChainerCV emphasizes modular design with a unified API such that users can compose the implementations in various methods.
The implementations include neural network models, data loaders, evaluation metrics, and visualization utilities.
The API is made consistent using the same data representation across the library.
For instance, we define the data representation for images, bounding boxes, semantic pixel-wise labels, instance mask, and key points.
Additionally, the API is consistent across similar functions.
For example, as mentioned previously, inference methods always take an iterable of images as inputs for all models.

In addition to rendering the interface intuitive, a unified API allows us to build utilities in it.
For example, we provide implementations that abstract the evaluation loop.
Internally, this abstraction iterates over the dataset, performs predictions from images, and calculates the evaluation metrics using the ground truth and the predictions.
It is noteworthy that the interface must be assumed for the data loader and the inference method such that the abstract utility can pass data among a data loader, an inference method, and an evaluation metric.
An example is shown in Figure~\ref{fig:chainercv_evaluation}.

\subsection{Reproducibility}
Reproducibility in machine learning and computer vision is an important factor affecting research quality.
ChainerCV aims at easing the process of reproducing the published results by providing a training code that is guaranteed to perform on par with them.
These algorithms would serve as baselines to obtain a new idea through refinement and as a tool to compare a new approach against the existing approaches.
With a careless implementation, the performance of trained models can easily change by deviating from the original logic and hyperparameters.
This type of mistakes disqualify the implementation as a useful baseline because researchers would not be able to attain competitive results and assess the impact of their ideas properly.
Table \ref{table:chainercv-models} shows the models supported by ChainerCV and the experimental results.
The reference scores are also presented, which are reported by the original papers or the authors' implementations.
As shown, the performance of our re-implementations is close to that of the original.
It is noteworthy that randomness included during training can be a reason for different scores.

\begin{figure}[!t]
  \begin{center}
    \begin{subfigure}{\linewidth}
      \begin{verbatim}dataset = VOCBboxDataset(year='2007', split='test')
it = SerialIterator(dataset, batch_size=2,
    repeat=False, shuffle=False)
model = SSD300(pretrained_model='voc0712')
evaluator = DetectionVOCEvaluator(it, model,
    label_names=voc_bbox_label_names)
# Run evaluation loop
result = evaluator()\end{verbatim}
    \end{subfigure}    
  \end{center}
  \caption{Example of performing an evaluation loop for object detection using \texttt{DetectionVOCEvaluator}.}
  \label{fig:chainercv_evaluation}
\end{figure}

\begin{table}[t]
  \caption{Supported models in ChainerCV and their scores for (1) image classification, (2) object detection, (3) semantic segmentation, and (4) instance segmentation. For image classification, we report the top 1 error.
For object detection, we report the mean average precision (mAP) for scores reported with Pascal VOC, and the average of mAP over different intersection over union threshold for scores with MS COCO.
For semantic segmentation, we report the mean intersection over union (mIoU).
For instance segmentation, we report the mAP of the mask.
}
  \label{table:chainercv-models}
  \begin{center}
    \begin{threeparttable}[t]
      {\tabcolsep=1.5mm
      \begin{tabular}{ccccc}
        \toprule
 	    \multirow{2}{*}{task} & \multirow{2}{*}{dataset} & \multirow{2}{*}{model} & \multicolumn{2}{c}{score} \\
        & & & reference & ours \\
        \midrule
        \multirow{9}{*}{(1)} & \multirow{9}{*}{ImageNet~\cite{imagenet}} & VGG16~\cite{Simonyan14c} & 28.5 & 29.0\tnote{*} \\
        & & ResNet50~\cite{He2016} & 24.7 & 24.8\tnote{*} \\
        & & ResNet101~\cite{He2016} & 23.6 & 23.6\tnote{*} \\
        & & ResNet152~\cite{He2016} & 23.0 & 23.2\tnote{*} \\
        & & SE-ResNet50~\cite{senet2018hu} & 22.4 & 22.7\tnote{*} \\
        & & SE-ResNet101~\cite{senet2018hu} & 21.8 & 21.8\tnote{*} \\
        & & SE-ResNet152~\cite{senet2018hu} & 21.3 & 21.4\tnote{*} \\
        & & SE-ResNeXt50~\cite{senet2018hu} & 21.0 & 20.9\tnote{*} \\
        & & SE-ResNeXt101~\cite{senet2018hu} & 19.8 & 19.7\tnote{*} \\
        \midrule
        \multirow{7}{*}{(2)} & \multirow{5}{*}{Pascal VOC~\cite{Everingham10}} & Faster R-CNN~\cite{Ren2015} & 73.2 & 74.7 \\
        & & SSD300~\cite{Liu2016} & 77.5 & 77.5 \\
        & & SSD512~\cite{Liu2016} & 79.5 & 79.7 \\
        & & YOLOv2~\cite{redmon2016yolo9000} & 75.8 & 75.8\tnote{*} \\
        & & YOLOv3~\cite{yolov3} & 80.2 & 80.2\tnote{*} \\ \cmidrule(l){2-5}
        & \multirow{2}{*}{MS COCO~\cite{lin2014microsoft}} & FPN ResNet50~\cite{lin2017feature} & 36.7 & 37.1 \\
        & & FPN ResNet101~\cite{lin2017feature} & 39.4 & 39.5\\
        \midrule
        \multirow{2}{*}{(3)} & CamVid~\cite{badrinarayanan2015segnet} & SegNet~\cite{badrinarayanan2015segnet} & 46.3 & 49.4 \\ \cmidrule(l){2-5}
        & CityScapes~\cite{cityscapes} & PSPNet~\cite{psp2017zhao} & 79.7 & 79.0\tnote{*} \\
        \midrule
        (4) & SBD~\cite{sbd2011bharath} & FCIS~\cite{fcis2017li} & 65.7 & 64.1\\
        \bottomrule
      \end{tabular}
      }
      \begin{tablenotes}
        \item[*]{We converted the weights of the original model}
      \end{tablenotes}
    \end{threeparttable}
  \end{center}
\end{table}

\section{Related Work}
To the best of our knowledge, Autograd~\cite{autograd} had adopted the Define-by-Run paradigm to construct the backward graph before it was proposed by Chainer.
Autograd is a library based on NumPy~\cite{oliphant_guide_2006} and designed to enable users to write a differentiable computational graph in Python code using NumPy.
However, it is not intended as a deep learning framework; therefore, it does not support GPU acceleration that is necessary for training deep models.
Thus, Chainer is the first framework that focuses on deep learning workloads with the Define-by-Run paradigm.
Currently, several other deep learning frameworks exist that adopt Define-by-Run.
PyTorch~\cite{pytorch} is a popular Define-by-Run framework inspired by Chainer~\footnote{https://github.com/pytorch/pytorch/blob/v0.4.1/README.md}, followed by Tensorflow~\cite{tensorflow2015-whitepaper} that introduced a feature called the "eager mode," which supports Define-by-Run model definitions.
MXNet~\cite{DBLP:journals/corr/ChenLLLWWXXZZ15} supports imperative tensor computations which can be combined with declarative symbolic expressions by a lazy evaluation.
PaddlePaddle~\cite{paddlepaddle} and CNTK~\cite{cntk} contains an imperative style as their optional programming style, while supporting the declarative style simultaneously.
In contrast to most of these frameworks that support Define-by-Run optionally, Chainer is highly optimized for Define-by-Run in its overall implementation and APIs.
This results in a simpler code base that reduces the cost for new developers to contribute to it.

DistBelief~\cite{DistBelief} first integrated three important techniques in the distributed training of deep
neural networks, data and model parallelism, asynchronous SGD, and
master-worker heterogeneous model.
Subsequently, these techniques became available in Tensorflow~\cite{tensorflow2015-whitepaper},
MXNet~\cite{DBLP:journals/corr/ChenLLLWWXXZZ15}, and
PaddlePaddle~\cite{paddlepaddle}.
However, asynchronous SGD cannot avoid stale gradients (\ref{syncvsasync}) that affect accuracy
and parameter server being bottleneck. To mitigate those issues, recent versions of them
have added options to run synchronous SGD~\cite{Xinghao2017, horovod}.

Meanwhile, Caffe2 and PyTorch~\cite{pytorch} use synchronous SGD.
To compensate for the cost of synchronization, Caffe2 and PyTorch minimized
the critical path of all-reduce communications through computation, by starting
all-reduce as soon as the backward computation of a layer
is completed.

Although open-sourcing models in computer vision is a widespread practice, we discovered a few studies that pursued a similar philosophy as that of ChainerCV.
Primary competitors include research program codes accompanied by papers.
Their primary purpose is to share research results in a verifiable manner; therefore, readability and modularity are often ignored.

Libraries more closely related to ChainerCV are \textit{pytorch/vision}~\cite{pytorch_vision} and GluonCV~\cite{gluoncv}.
\textit{pytorch/vision} is a computer vision library that uses PyTorch as its backend.
At the time of writing, its support for pretrained models is limited only to image classification.
GluonCV is a recently released computer vision library that uses Gluon~\cite{gluon}, which is another deep learning framework, as its backend.
Similar to ChainerCV, GluonCV supports object detection, semantic segmentation, and instance segmentation.
However, they do not pursue reproducibility as their core goal.

\section{Conclusion}


This paper introduced Chainer, a deep learning framework that enabled users to easily implement new algorithms and complex neural networks.  
Chainer has already been used successfully in a variety of leading-edge applications, including deep reinforcement learning~\cite{mnih2013playing}, word2vec distributed representations~\cite{word2vec}, recurrent neural network language models~\cite{rnnlm}, human pose estimation~\cite{Toshev_2014_CVPR}, and variational auto-encoders~\cite{vae}.
Because dedicated developers and users worldwide are actively collaborating on GitHub to improve Chainer, we anticipate that Chainer will become more versatile and useful in the future.
In particular, the performance improvement attained by improving CuPy allows users to apply various types of deep learning models with CPU/GPU-agnostic codes.
We invite all members of the deep learning community to test out Chainer and to contribute to its development.

%
\begin{acks}
This work could not be achieved without the help of all the contributors and the feedback from users of Chainer, CuPy, ChainerCV, and related projects. We express special thanks to them.
\end{acks}

%
\bibliographystyle{ACM-Reference-Format}
\bibliography{main}


\begin{thebibliography}{52}


\ifx \showCODEN    \undefined \def \showCODEN     #1{\unskip}     \fi
\ifx \showDOI      \undefined \def \showDOI       #1{#1}\fi
\ifx \showISBNx    \undefined \def \showISBNx     #1{\unskip}     \fi
\ifx \showISBNxiii \undefined \def \showISBNxiii  #1{\unskip}     \fi
\ifx \showISSN     \undefined \def \showISSN      #1{\unskip}     \fi
\ifx \showLCCN     \undefined \def \showLCCN      #1{\unskip}     \fi
\ifx \shownote     \undefined \def \shownote      #1{#1}          \fi
\ifx \showarticletitle \undefined \def \showarticletitle #1{#1}   \fi
\ifx \showURL      \undefined \def \showURL       {\relax}        \fi
\providecommand\bibfield[2]{#2}
\providecommand\bibinfo[2]{#2}
\providecommand\natexlab[1]{#1}
\providecommand\showeprint[2][]{arXiv:#2}

\bibitem[\protect\citeauthoryear{??}{glu}{[n. d.]a}]%
        {gluon}
 \bibinfo{year}{[n. d.]}\natexlab{a}.
\newblock \bibinfo{title}{Gluon}.
\newblock \bibinfo{howpublished}{\url{https://gluon.mxnet.io}}.
\newblock


\bibitem[\protect\citeauthoryear{??}{glu}{[n. d.]b}]%
        {gluoncv}
 \bibinfo{year}{[n. d.]}\natexlab{b}.
\newblock \bibinfo{title}{GluonCV}.
\newblock \bibinfo{howpublished}{\url{https://gluon-cv.mxnet.io/}}.
\newblock


\bibitem[\protect\citeauthoryear{??}{pad}{[n. d.]}]%
        {paddlepaddle}
 \bibinfo{year}{[n. d.]}\natexlab{}.
\newblock \bibinfo{title}{PaddlePaddle}.
\newblock \bibinfo{howpublished}{\url{http://www.paddlepaddle.org/}}.
\newblock


\bibitem[\protect\citeauthoryear{??}{pyt}{[n. d.]}]%
        {pytorch_vision}
 \bibinfo{year}{[n. d.]}\natexlab{}.
\newblock \bibinfo{title}{pytorch/vision}.
\newblock \bibinfo{howpublished}{\url{https://github.com/pytorch/vision}}.
\newblock


\bibitem[\protect\citeauthoryear{??}{ncc}{2017}]%
        {nccl}
 \bibinfo{year}{2017}\natexlab{}.
\newblock \bibinfo{title}{{NVIDIA Collective Communications Library (NCCL)}}.
\newblock \bibinfo{howpublished}{\url{https://developer.nvidia.com/nccl}}.
\newblock


\bibitem[\protect\citeauthoryear{Abadi, Agarwal, Barham, Brevdo, Chen, Citro,
  Corrado, Davis, Dean, Devin, Ghemawat, Goodfellow, Harp, Irving, Isard, Jia,
  Jozefowicz, Kaiser, Kudlur, Levenberg, Man\'{e}, Monga, Moore, Murray, Olah,
  Schuster, Shlens, Steiner, Sutskever, Talwar, Tucker, Vanhoucke, Vasudevan,
  Vi\'{e}gas, Vinyals, Warden, Wattenberg, Wicke, Yu, and Zheng}{Abadi
  et~al\mbox{.}}{2015}]%
        {tensorflow2015-whitepaper}
\bibfield{author}{\bibinfo{person}{Mart\'{\i}n Abadi}, \bibinfo{person}{Ashish
  Agarwal}, \bibinfo{person}{Paul Barham}, \bibinfo{person}{Eugene Brevdo},
  \bibinfo{person}{Zhifeng Chen}, \bibinfo{person}{Craig Citro},
  \bibinfo{person}{Greg~S. Corrado}, \bibinfo{person}{Andy Davis},
  \bibinfo{person}{Jeffrey Dean}, \bibinfo{person}{Matthieu Devin},
  \bibinfo{person}{Sanjay Ghemawat}, \bibinfo{person}{Ian Goodfellow},
  \bibinfo{person}{Andrew Harp}, \bibinfo{person}{Geoffrey Irving},
  \bibinfo{person}{Michael Isard}, \bibinfo{person}{Yangqing Jia},
  \bibinfo{person}{Rafal Jozefowicz}, \bibinfo{person}{Lukasz Kaiser},
  \bibinfo{person}{Manjunath Kudlur}, \bibinfo{person}{Josh Levenberg},
  \bibinfo{person}{Dan Man\'{e}}, \bibinfo{person}{Rajat Monga},
  \bibinfo{person}{Sherry Moore}, \bibinfo{person}{Derek Murray},
  \bibinfo{person}{Chris Olah}, \bibinfo{person}{Mike Schuster},
  \bibinfo{person}{Jonathon Shlens}, \bibinfo{person}{Benoit Steiner},
  \bibinfo{person}{Ilya Sutskever}, \bibinfo{person}{Kunal Talwar},
  \bibinfo{person}{Paul Tucker}, \bibinfo{person}{Vincent Vanhoucke},
  \bibinfo{person}{Vijay Vasudevan}, \bibinfo{person}{Fernanda Vi\'{e}gas},
  \bibinfo{person}{Oriol Vinyals}, \bibinfo{person}{Pete Warden},
  \bibinfo{person}{Martin Wattenberg}, \bibinfo{person}{Martin Wicke},
  \bibinfo{person}{Yuan Yu}, {and} \bibinfo{person}{Xiaoqiang Zheng}.}
  \bibinfo{year}{2015}\natexlab{}.
\newblock \bibinfo{title}{{TensorFlow}: Large-Scale Machine Learning on
  Heterogeneous Systems}.
\newblock
\newblock
\urldef\tempurl%
\url{http://tensorflow.org/}
\showURL{%
\tempurl}
\newblock
\shownote{Software available from tensorflow.org.}


\bibitem[\protect\citeauthoryear{Adam~Paszke and Chanan}{Adam~Paszke and
  Chanan}{[n. d.]}]%
        {pytorch}
\bibfield{author}{\bibinfo{person}{Soumith~Chintala Adam~Paszke, Sam~Gross}
  {and} \bibinfo{person}{Gregory Chanan}.} \bibinfo{year}{[n. d.]}\natexlab{}.
\newblock \bibinfo{title}{PyTorch}.
\newblock
\newblock
\newblock
\shownote{https://github.com/pytorch/pytorch.}


\bibitem[\protect\citeauthoryear{Badrinarayanan, Kendall, and
  Cipolla}{Badrinarayanan et~al\mbox{.}}{2017}]%
        {badrinarayanan2015segnet}
\bibfield{author}{\bibinfo{person}{Vijay Badrinarayanan}, \bibinfo{person}{Alex
  Kendall}, {and} \bibinfo{person}{Roberto Cipolla}.}
  \bibinfo{year}{2017}\natexlab{}.
\newblock \showarticletitle{SegNet: A Deep Convolutional Encoder-Decoder
  Architecture for Image Segmentation}.
\newblock \bibinfo{journal}{\emph{IEEE Transactions on Pattern Analysis and
  Machine Intelligence}} (\bibinfo{year}{2017}).
\newblock


\bibitem[\protect\citeauthoryear{Chen, Li, Li, Lin, Wang, Wang, Xiao, Xu,
  Zhang, and Zhang}{Chen et~al\mbox{.}}{2015}]%
        {DBLP:journals/corr/ChenLLLWWXXZZ15}
\bibfield{author}{\bibinfo{person}{Tianqi Chen}, \bibinfo{person}{Mu Li},
  \bibinfo{person}{Yutian Li}, \bibinfo{person}{Min Lin},
  \bibinfo{person}{Naiyan Wang}, \bibinfo{person}{Minjie Wang},
  \bibinfo{person}{Tianjun Xiao}, \bibinfo{person}{Bing Xu},
  \bibinfo{person}{Chiyuan Zhang}, {and} \bibinfo{person}{Zheng Zhang}.}
  \bibinfo{year}{2015}\natexlab{}.
\newblock \showarticletitle{MXNet: {A} Flexible and Efficient Machine Learning
  Library for Heterogeneous Distributed Systems}.
\newblock \bibinfo{journal}{\emph{CoRR}}  \bibinfo{volume}{abs/1512.01274}
  (\bibinfo{year}{2015}).
\newblock
\showeprint[arxiv]{1512.01274}


\bibitem[\protect\citeauthoryear{Codreanu, Podareanu, and Saletore}{Codreanu
  et~al\mbox{.}}{2017}]%
        {Codreanu2017}
\bibfield{author}{\bibinfo{person}{Valeriu Codreanu}, \bibinfo{person}{Damian
  Podareanu}, {and} \bibinfo{person}{Vikram Saletore}.}
  \bibinfo{year}{2017}\natexlab{}.
\newblock \bibinfo{title}{Achieving Deep Learning Training in less than 40
  Minutes on ImageNet-1K}.
\newblock
  \bibinfo{howpublished}{\url{https://blog.surf.nl/en/imagenet-1k-training-on-intel-xeon-\\phi-in-less-than-40-minutes/}}.
\newblock


\bibitem[\protect\citeauthoryear{Collobert}{Collobert}{2008}]%
        {collobert:2008a}
\bibfield{author}{\bibinfo{person}{R. Collobert}.}
  \bibinfo{year}{2008}\natexlab{}.
\newblock \bibinfo{title}{Torch}.
\newblock \bibinfo{howpublished}{NIPS Workshop on Machine Learning Open Source
  Software}.
\newblock


\bibitem[\protect\citeauthoryear{David~Oro and Hernando}{David~Oro and
  Hernando}{2016}]%
        {nms}
\bibfield{author}{\bibinfo{person}{Xavier~Martorell David~Oro,
  Carles~Fernandez} {and} \bibinfo{person}{Javier Hernando}.}
  \bibinfo{year}{2016}\natexlab{}.
\newblock \showarticletitle{Work-Efficient Parallel non-maximum suppression for
  embedded GPU architecture}.
\newblock \bibinfo{journal}{\emph{ICASSP}} (\bibinfo{year}{2016}).
\newblock


\bibitem[\protect\citeauthoryear{Dean, Corrado, Monga, Chen, Devin, Mao,
  aurelio Ranzato, Senior, Tucker, Yang, Le, and Ng}{Dean
  et~al\mbox{.}}{2012}]%
        {DistBelief}
\bibfield{author}{\bibinfo{person}{Jeffrey Dean}, \bibinfo{person}{Greg
  Corrado}, \bibinfo{person}{Rajat Monga}, \bibinfo{person}{Kai Chen},
  \bibinfo{person}{Matthieu Devin}, \bibinfo{person}{Mark Mao},
  \bibinfo{person}{Marc\textquotesingle aurelio Ranzato},
  \bibinfo{person}{Andrew Senior}, \bibinfo{person}{Paul Tucker},
  \bibinfo{person}{Ke Yang}, \bibinfo{person}{Quoc~V. Le}, {and}
  \bibinfo{person}{Andrew~Y. Ng}.} \bibinfo{year}{2012}\natexlab{}.
\newblock \showarticletitle{Large Scale Distributed Deep Networks}.
\newblock In \bibinfo{booktitle}{\emph{Advances in Neural Information
  Processing Systems 25}}, \bibfield{editor}{\bibinfo{person}{F.~Pereira},
  \bibinfo{person}{C.~J.~C. Burges}, \bibinfo{person}{L.~Bottou}, {and}
  \bibinfo{person}{K.~Q. Weinberger}} (Eds.). \bibinfo{publisher}{Curran
  Associates, Inc.}, \bibinfo{pages}{1223--1231}.
\newblock


\bibitem[\protect\citeauthoryear{Dean and Ghemawat}{Dean and Ghemawat}{2004}]%
        {dean2004}
\bibfield{author}{\bibinfo{person}{Jeffrey Dean} {and} \bibinfo{person}{Sanjay
  Ghemawat}.} \bibinfo{year}{2004}\natexlab{}.
\newblock \showarticletitle{MapReduce: Simplified Data Processing on Large
  Clusters}, In \bibinfo{booktitle}{OSDI 2004}.
\newblock \bibinfo{journal}{\emph{OSDI '04}}, \bibinfo{pages}{137--150}.
\newblock


\bibitem[\protect\citeauthoryear{Deng, Dong, Socher, Li, Li, and Fei-Fei}{Deng
  et~al\mbox{.}}{2009}]%
        {imagenet2009}
\bibfield{author}{\bibinfo{person}{J. Deng}, \bibinfo{person}{W. Dong},
  \bibinfo{person}{R. Socher}, \bibinfo{person}{L.-J. Li}, \bibinfo{person}{K.
  Li}, {and} \bibinfo{person}{L. Fei-Fei}.} \bibinfo{year}{2009}\natexlab{}.
\newblock \showarticletitle{{ImageNet: A Large-Scale Hierarchical Image
  Database}}. In \bibinfo{booktitle}{\emph{CVPR09}}.
\newblock


\bibitem[\protect\citeauthoryear{Dougal~Maclaurin}{Dougal~Maclaurin}{[n. d.]}]%
        {autograd}
\bibfield{author}{\bibinfo{person}{et.~al. Dougal~Maclaurin}.}
  \bibinfo{year}{[n. d.]}\natexlab{}.
\newblock \bibinfo{title}{Autograd}.
\newblock
\newblock
\urldef\tempurl%
\url{https://github.com/HIPS/autograd}
\showURL{%
\tempurl}


\bibitem[\protect\citeauthoryear{Everingham, Van~Gool, Williams, Winn, and
  Zisserman}{Everingham et~al\mbox{.}}{2010}]%
        {Everingham10}
\bibfield{author}{\bibinfo{person}{M. Everingham}, \bibinfo{person}{L.
  Van~Gool}, \bibinfo{person}{C.~K.~I. Williams}, \bibinfo{person}{J. Winn},
  {and} \bibinfo{person}{A. Zisserman}.} \bibinfo{year}{2010}\natexlab{}.
\newblock \showarticletitle{The Pascal Visual Object Classes (VOC) Challenge}.
\newblock \bibinfo{journal}{\emph{IJCV}} \bibinfo{volume}{88},
  \bibinfo{number}{2} (\bibinfo{date}{June} \bibinfo{year}{2010}),
  \bibinfo{pages}{303--338}.
\newblock


\bibitem[\protect\citeauthoryear{Goodfellow, Warde-Farley, Lamblin, Dumoulin,
  Mirza, Pascanu, Bergstra, Bastien, and Bengio}{Goodfellow
  et~al\mbox{.}}{2013}]%
        {journals/corr/GoodfellowWLDMPBBB13}
\bibfield{author}{\bibinfo{person}{Ian~J. Goodfellow}, \bibinfo{person}{David
  Warde-Farley}, \bibinfo{person}{Pascal Lamblin}, \bibinfo{person}{Vincent
  Dumoulin}, \bibinfo{person}{Mehdi Mirza}, \bibinfo{person}{Razvan Pascanu},
  \bibinfo{person}{James Bergstra}, \bibinfo{person}{Fr辿d辿ric Bastien},
  {and} \bibinfo{person}{Yoshua Bengio}.} \bibinfo{year}{2013}\natexlab{}.
\newblock \showarticletitle{Pylearn2: a machine learning research library.}
\newblock \bibinfo{journal}{\emph{CoRR}}  \bibinfo{volume}{abs/1308.4214}
  (\bibinfo{year}{2013}).
\newblock


\bibitem[\protect\citeauthoryear{Goyal, Doll{\'{a}}r, Girshick, Noordhuis,
  Wesolowski, Kyrola, Tulloch, Jia, and He}{Goyal et~al\mbox{.}}{2017}]%
        {Goyal2017}
\bibfield{author}{\bibinfo{person}{Priya Goyal}, \bibinfo{person}{Piotr
  Doll{\'{a}}r}, \bibinfo{person}{Ross~B. Girshick}, \bibinfo{person}{Pieter
  Noordhuis}, \bibinfo{person}{Lukasz Wesolowski}, \bibinfo{person}{Aapo
  Kyrola}, \bibinfo{person}{Andrew Tulloch}, \bibinfo{person}{Yangqing Jia},
  {and} \bibinfo{person}{Kaiming He}.} \bibinfo{year}{2017}\natexlab{}.
\newblock \showarticletitle{Accurate, Large Minibatch {SGD:} Training
  {ImageNet} in 1 Hour}.
\newblock \bibinfo{journal}{\emph{CoRR}}  \bibinfo{volume}{abs/1706.02677}
  (\bibinfo{year}{2017}).
\newblock


\bibitem[\protect\citeauthoryear{Hariharan, Arbelaez, Bourdev, Maji, and
  Malik}{Hariharan et~al\mbox{.}}{2011}]%
        {sbd2011bharath}
\bibfield{author}{\bibinfo{person}{Bharath Hariharan}, \bibinfo{person}{Pablo
  Arbelaez}, \bibinfo{person}{Lubomir Bourdev}, \bibinfo{person}{Subhransu
  Maji}, {and} \bibinfo{person}{Jitendra Malik}.}
  \bibinfo{year}{2011}\natexlab{}.
\newblock \showarticletitle{Semantic Contours from Inverse Detectors}. In
  \bibinfo{booktitle}{\emph{ICCV}}.
\newblock


\bibitem[\protect\citeauthoryear{He, Zhang, Ren, and Sun}{He
  et~al\mbox{.}}{2016}]%
        {He2016}
\bibfield{author}{\bibinfo{person}{Kaiming He}, \bibinfo{person}{Xiangyu
  Zhang}, \bibinfo{person}{Shaoqing Ren}, {and} \bibinfo{person}{Jian Sun}.}
  \bibinfo{year}{2016}\natexlab{}.
\newblock \showarticletitle{Deep Residual Learning for Image Recognition}. In
  \bibinfo{booktitle}{\emph{CVPR}}. \bibinfo{pages}{770--778}.
\newblock


\bibitem[\protect\citeauthoryear{Hengshuang~Zhao}{Hengshuang~Zhao}{2017}]%
        {psp2017zhao}
\bibfield{author}{\bibinfo{person}{Xiaojuan Qi Xiaogang Wang Jiaya~Jia
  Hengshuang~Zhao, Jianping~Shi}.} \bibinfo{year}{2017}\natexlab{}.
\newblock \showarticletitle{Pyramid Scene Parsing Network}.
\newblock \bibinfo{journal}{\emph{CVPR}} (\bibinfo{year}{2017}).
\newblock


\bibitem[\protect\citeauthoryear{Honnibal and Montani}{Honnibal and
  Montani}{[n. d.]}]%
        {spaCy}
\bibfield{author}{\bibinfo{person}{Matthew Honnibal} {and}
  \bibinfo{person}{Ines Montani}.} \bibinfo{year}{[n. d.]}\natexlab{}.
\newblock \showarticletitle{spaCy 2: Natural language understanding with Bloom
  embeddings, convolutional neural networks and incremental parsing}.
\newblock  (\bibinfo{year}{[n. d.]}).
\newblock
\urldef\tempurl%
\url{https://spacy.io/}
\showURL{%
\tempurl}


\bibitem[\protect\citeauthoryear{Hu, Shen, and Sun}{Hu et~al\mbox{.}}{2018}]%
        {senet2018hu}
\bibfield{author}{\bibinfo{person}{Jie Hu}, \bibinfo{person}{Li Shen}, {and}
  \bibinfo{person}{Gang Sun}.} \bibinfo{year}{2018}\natexlab{}.
\newblock \showarticletitle{Squeeze-and-Excitation Networks}.
\newblock \bibinfo{journal}{\emph{CVPR}}.
\newblock


\bibitem[\protect\citeauthoryear{Jia}{Jia}{2013}]%
        {Jia13caffe}
\bibfield{author}{\bibinfo{person}{Yangqing Jia}.}
  \bibinfo{year}{2013}\natexlab{}.
\newblock \bibinfo{title}{{Caffe}: An Open Source Convolutional Architecture
  for Fast Feature Embedding}.
\newblock
\newblock


\bibitem[\protect\citeauthoryear{Jones, Oliphant, Peterson,
  et~al\mbox{.}}{Jones et~al\mbox{.}}{01  }]%
        {jones2001scipy}
\bibfield{author}{\bibinfo{person}{Eric Jones}, \bibinfo{person}{Travis
  Oliphant}, \bibinfo{person}{Pearu Peterson}, {et~al\mbox{.}}}
  \bibinfo{year}{2001--}\natexlab{}.
\newblock \bibinfo{title}{{SciPy}: Open source scientific tools for {Python}}.
\newblock
\newblock
\urldef\tempurl%
\url{http://www.scipy.org/}
\showURL{%
\tempurl}


\bibitem[\protect\citeauthoryear{Kingma and Welling}{Kingma and
  Welling}{2014}]%
        {vae}
\bibfield{author}{\bibinfo{person}{Diederik~P. Kingma} {and}
  \bibinfo{person}{Max Welling}.} \bibinfo{year}{2014}\natexlab{}.
\newblock \showarticletitle{Auto-Encoding Variational Bayes}.
\newblock \bibinfo{journal}{\emph{ICLR}} (\bibinfo{year}{2014}).
\newblock


\bibitem[\protect\citeauthoryear{{Kl{\"o}ckner}, {Pinto}, {Lee}, {Catanzaro},
  {Ivanov}, and {Fasih}}{{Kl{\"o}ckner} et~al\mbox{.}}{2012}]%
        {pycuda}
\bibfield{author}{\bibinfo{person}{Andreas {Kl{\"o}ckner}},
  \bibinfo{person}{Nicolas {Pinto}}, \bibinfo{person}{Yunsup {Lee}},
  \bibinfo{person}{B. {Catanzaro}}, \bibinfo{person}{Paul {Ivanov}}, {and}
  \bibinfo{person}{Ahmed {Fasih}}.} \bibinfo{year}{2012}\natexlab{}.
\newblock \showarticletitle{{PyCUDA and PyOpenCL: A Scripting-Based Approach to
  GPU Run-Time Code Generation}}.
\newblock \bibinfo{journal}{\emph{Parallel Comput.}} \bibinfo{volume}{38},
  \bibinfo{number}{3} (\bibinfo{year}{2012}), \bibinfo{pages}{157--174}.
\newblock
\showISSN{0167-8191}
\urldef\tempurl%
\url{https://doi.org/10.1016/j.parco.2011.09.001}
\showDOI{\tempurl}


\bibitem[\protect\citeauthoryear{LeCun, Bengio, and Hinton}{LeCun
  et~al\mbox{.}}{2015}]%
        {mafia}
\bibfield{author}{\bibinfo{person}{Yann LeCun}, \bibinfo{person}{Yoshua
  Bengio}, {and} \bibinfo{person}{Geoffrey Hinton}.}
  \bibinfo{year}{2015}\natexlab{}.
\newblock \showarticletitle{Deep learning}.
\newblock \bibinfo{journal}{\emph{Nature}}  \bibinfo{volume}{521}
  (\bibinfo{year}{2015}), \bibinfo{pages}{436--444}.
\newblock


\bibitem[\protect\citeauthoryear{Lin, Doll{\'a}r, Girshick, He, Hariharan, and
  Belongie}{Lin et~al\mbox{.}}{2017}]%
        {lin2017feature}
\bibfield{author}{\bibinfo{person}{Tsung-Yi Lin}, \bibinfo{person}{Piotr
  Doll{\'a}r}, \bibinfo{person}{Ross~B Girshick}, \bibinfo{person}{Kaiming He},
  \bibinfo{person}{Bharath Hariharan}, {and} \bibinfo{person}{Serge~J
  Belongie}.} \bibinfo{year}{2017}\natexlab{}.
\newblock \showarticletitle{Feature Pyramid Networks for Object Detection.}. In
  \bibinfo{booktitle}{\emph{CVPR}}, Vol.~\bibinfo{volume}{1}.
  \bibinfo{pages}{3}.
\newblock


\bibitem[\protect\citeauthoryear{Lin, Maire, Belongie, Hays, Perona, Ramanan,
  Doll{\'a}r, and Zitnick}{Lin et~al\mbox{.}}{2014}]%
        {lin2014microsoft}
\bibfield{author}{\bibinfo{person}{Tsung-Yi Lin}, \bibinfo{person}{Michael
  Maire}, \bibinfo{person}{Serge Belongie}, \bibinfo{person}{James Hays},
  \bibinfo{person}{Pietro Perona}, \bibinfo{person}{Deva Ramanan},
  \bibinfo{person}{Piotr Doll{\'a}r}, {and} \bibinfo{person}{C~Lawrence
  Zitnick}.} \bibinfo{year}{2014}\natexlab{}.
\newblock \showarticletitle{Microsoft coco: Common objects in context}. In
  \bibinfo{booktitle}{\emph{European conference on computer vision}}. Springer,
  \bibinfo{pages}{740--755}.
\newblock


\bibitem[\protect\citeauthoryear{Liu, Anguelov, Erhan, Szegedy, Reed, Fu, and
  Berg}{Liu et~al\mbox{.}}{2016}]%
        {Liu2016}
\bibfield{author}{\bibinfo{person}{Wei Liu}, \bibinfo{person}{Dragomir
  Anguelov}, \bibinfo{person}{Dumitru Erhan}, \bibinfo{person}{Christian
  Szegedy}, \bibinfo{person}{Scott Reed}, \bibinfo{person}{Cheng-yang Fu},
  {and} \bibinfo{person}{Alexander~C Berg}.} \bibinfo{year}{2016}\natexlab{}.
\newblock \showarticletitle{{SSD: Single Shot MultiBox Detector}}.
\newblock \bibinfo{journal}{\emph{arXiv preprint arXiv:1512.02325v2}}
  (\bibinfo{year}{2016}).
\newblock


\bibitem[\protect\citeauthoryear{Marius~Cordts}{Marius~Cordts}{2017}]%
        {cityscapes}
\bibfield{author}{\bibinfo{person}{Sebastian Ramos Timo Rehfeld Markus
  Enzweiler Rodrigo Benenson Uwe Franke Stefan Roth Bernt~Schiele
  Marius~Cordts, Mohamed~Omran}.} \bibinfo{year}{2017}\natexlab{}.
\newblock \showarticletitle{The Cityscapes Dataset for Semantic Urban Scene
  Understanding}.
\newblock \bibinfo{journal}{\emph{CVPR}} (\bibinfo{year}{2017}).
\newblock


\bibitem[\protect\citeauthoryear{Mikolov, Karafi{\'{a}}t, Burget,
  {\v{C}}ernock{\'{y}}, and Khudanpur}{Mikolov et~al\mbox{.}}{2010}]%
        {rnnlm}
\bibfield{author}{\bibinfo{person}{Tom{\'{a}}{\v{s}} Mikolov},
  \bibinfo{person}{Martin Karafi{\'{a}}t}, \bibinfo{person}{Luk{\'{a}}{\v{s}}
  Burget}, \bibinfo{person}{Jan {\v{C}}ernock{\'{y}}}, {and}
  \bibinfo{person}{Sanjeev Khudanpur}.} \bibinfo{year}{2010}\natexlab{}.
\newblock \showarticletitle{Recurrent neural network based language model}. In
  \bibinfo{booktitle}{\emph{INTERSPEECH}}. \bibinfo{pages}{1045--1048}.
\newblock


\bibitem[\protect\citeauthoryear{Mikolov, Sutskever, Chen, Corrado, and
  Dean}{Mikolov et~al\mbox{.}}{2013}]%
        {word2vec}
\bibfield{author}{\bibinfo{person}{Tomas Mikolov}, \bibinfo{person}{Ilya
  Sutskever}, \bibinfo{person}{Kai Chen}, \bibinfo{person}{Greg~S Corrado},
  {and} \bibinfo{person}{Jeff Dean}.} \bibinfo{year}{2013}\natexlab{}.
\newblock \showarticletitle{Distributed Representations of Words and Phrases
  and their Compositionality}.
\newblock \bibinfo{journal}{\emph{NIPS}} (\bibinfo{year}{2013}),
  \bibinfo{pages}{3111--3119}.
\newblock


\bibitem[\protect\citeauthoryear{Mnih, Kavukcuoglu, Silver, Graves, Antonoglou,
  Wierstra, and Riedmiller}{Mnih et~al\mbox{.}}{2013}]%
        {mnih2013playing}
\bibfield{author}{\bibinfo{person}{Volodymyr Mnih}, \bibinfo{person}{Koray
  Kavukcuoglu}, \bibinfo{person}{David Silver}, \bibinfo{person}{Alex Graves},
  \bibinfo{person}{Ioannis Antonoglou}, \bibinfo{person}{Daan Wierstra}, {and}
  \bibinfo{person}{Martin Riedmiller}.} \bibinfo{year}{2013}\natexlab{}.
\newblock \bibinfo{title}{Playing Atari with Deep Reinforcement Learning}.
\newblock
\newblock
\newblock
\shownote{NIPS Deep Learning Workshop.}


\bibitem[\protect\citeauthoryear{Olga~Russakovsky}{Olga~Russakovsky}{2015}]%
        {imagenet}
\bibfield{author}{\bibinfo{person}{Hao Su Jonathan Krause Sanjeev Satheesh Sean
  Ma Zhiheng Huang Andrej Karpathy Aditya Khosla Michael Bernstein Alexander C.
  Berg Li Fei-Fei Olga~Russakovsky, Jia~Deng}.}
  \bibinfo{year}{2015}\natexlab{}.
\newblock \showarticletitle{ImageNet Large Scale Visual Recognition Challenge}.
\newblock \bibinfo{journal}{\emph{IJCV}} (\bibinfo{year}{2015}).
\newblock


\bibitem[\protect\citeauthoryear{Oliphant}{Oliphant}{2006}]%
        {oliphant_guide_2006}
\bibfield{author}{\bibinfo{person}{Travis Oliphant}.}
  \bibinfo{year}{2006}\natexlab{}.
\newblock \bibinfo{booktitle}{\emph{Guide to {NumPy}}}.
\newblock \bibinfo{publisher}{Trelgol Publishing}.
\newblock
\urldef\tempurl%
\url{http://www.tramy.us/numpybook.pdf}
\showURL{%
\tempurl}


\bibitem[\protect\citeauthoryear{Pan, Chen, Monga, Bengio, and Jozefowicz}{Pan
  et~al\mbox{.}}{2017}]%
        {Xinghao2017}
\bibfield{author}{\bibinfo{person}{Xinghao Pan}, \bibinfo{person}{Jianmin
  Chen}, \bibinfo{person}{Rajat Monga}, \bibinfo{person}{Samy Bengio}, {and}
  \bibinfo{person}{Rafal Jozefowicz}.} \bibinfo{year}{2017}\natexlab{}.
\newblock \showarticletitle{Revisiting Distributed Synchronous SGD}.
\newblock \bibinfo{journal}{\emph{ICLR Workshop Track, 2016}}
  (\bibinfo{date}{02} \bibinfo{year}{2017}).
\newblock


\bibitem[\protect\citeauthoryear{Redmon and Farhadi}{Redmon and
  Farhadi}{2016}]%
        {redmon2016yolo9000}
\bibfield{author}{\bibinfo{person}{Joseph Redmon} {and} \bibinfo{person}{Ali
  Farhadi}.} \bibinfo{year}{2016}\natexlab{}.
\newblock \showarticletitle{YOLO9000: Better, Faster, Stronger}.
\newblock \bibinfo{journal}{\emph{arXiv preprint arXiv:1612.08242}}
  (\bibinfo{year}{2016}).
\newblock


\bibitem[\protect\citeauthoryear{Redmon and Farhadi}{Redmon and
  Farhadi}{2018}]%
        {yolov3}
\bibfield{author}{\bibinfo{person}{Joseph Redmon} {and} \bibinfo{person}{Ali
  Farhadi}.} \bibinfo{year}{2018}\natexlab{}.
\newblock \showarticletitle{YOLOv3: An Incremental Improvement}.
\newblock \bibinfo{journal}{\emph{arXiv}} (\bibinfo{year}{2018}).
\newblock


\bibitem[\protect\citeauthoryear{Ren, He, Girshick, and Sun}{Ren
  et~al\mbox{.}}{2015}]%
        {Ren2015}
\bibfield{author}{\bibinfo{person}{Shaoqing Ren}, \bibinfo{person}{Kaiming He},
  \bibinfo{person}{Ross Girshick}, {and} \bibinfo{person}{Jian Sun}.}
  \bibinfo{year}{2015}\natexlab{}.
\newblock \showarticletitle{Faster R-CNN: Towards Real-Time Object Detection
  with Region Proposal Networks}.
\newblock In \bibinfo{booktitle}{\emph{Advances in Neural Information
  Processing Systems 28}}, \bibfield{editor}{\bibinfo{person}{C.~Cortes},
  \bibinfo{person}{N.~D. Lawrence}, \bibinfo{person}{D.~D. Lee},
  \bibinfo{person}{M.~Sugiyama}, {and} \bibinfo{person}{R.~Garnett}} (Eds.).
  \bibinfo{publisher}{Curran Associates, Inc.}, \bibinfo{pages}{91--99}.
\newblock


\bibitem[\protect\citeauthoryear{Schreiber}{Schreiber}{2017}]%
        {pomegranate2017}
\bibfield{author}{\bibinfo{person}{Jacob Schreiber}.}
  \bibinfo{year}{2017}\natexlab{}.
\newblock \showarticletitle{Pomegranate: fast and flexible probabilistic
  modeling in python.}
\newblock \bibinfo{journal}{\emph{CoRR}}  \bibinfo{volume}{abs/1711.00137}
  (\bibinfo{year}{2017}).
\newblock


\bibitem[\protect\citeauthoryear{Sergeev and Balso}{Sergeev and Balso}{2018}]%
        {horovod}
\bibfield{author}{\bibinfo{person}{Alexander Sergeev} {and}
  \bibinfo{person}{Mike~Del Balso}.} \bibinfo{year}{2018}\natexlab{}.
\newblock \showarticletitle{Horovod: fast and easy distributed deep learning in
  TensorFlow}.
\newblock \bibinfo{journal}{\emph{CoRR}}  \bibinfo{volume}{abs/1802.05799}
  (\bibinfo{year}{2018}).
\newblock
\showeprint[arxiv]{1802.05799}
\urldef\tempurl%
\url{http://arxiv.org/abs/1802.05799}
\showURL{%
\tempurl}


\bibitem[\protect\citeauthoryear{Shazeer, Cheng, Parmar, Tran, Vaswani,
  Koanantakool, Hawkins, Lee, Hong, Young, Sepassi, and Hechtman}{Shazeer
  et~al\mbox{.}}{2018}]%
        {shazeer2018mesh}
\bibfield{author}{\bibinfo{person}{Noam Shazeer}, \bibinfo{person}{Youlong
  Cheng}, \bibinfo{person}{Niki Parmar}, \bibinfo{person}{Dustin Tran},
  \bibinfo{person}{Ashish Vaswani}, \bibinfo{person}{Penporn Koanantakool},
  \bibinfo{person}{Peter Hawkins}, \bibinfo{person}{HyoukJoong Lee},
  \bibinfo{person}{Mingsheng Hong}, \bibinfo{person}{Cliff Young},
  \bibinfo{person}{Ryan Sepassi}, {and} \bibinfo{person}{Blake Hechtman}.}
  \bibinfo{year}{2018}\natexlab{}.
\newblock \showarticletitle{{Mesh-TensorFlow}: Deep Learning for
  Supercomputers}. In \bibinfo{booktitle}{\emph{Neural Information Processing
  Systems}}.
\newblock


\bibitem[\protect\citeauthoryear{Simonyan and Zisserman}{Simonyan and
  Zisserman}{2014}]%
        {Simonyan14c}
\bibfield{author}{\bibinfo{person}{K. Simonyan} {and} \bibinfo{person}{A.
  Zisserman}.} \bibinfo{year}{2014}\natexlab{}.
\newblock \showarticletitle{Very Deep Convolutional Networks for Large-Scale
  Image Recognition}.
\newblock \bibinfo{journal}{\emph{CoRR}}  \bibinfo{volume}{abs/1409.1556}
  (\bibinfo{year}{2014}).
\newblock


\bibitem[\protect\citeauthoryear{Sutskever, Vinyals, and Le}{Sutskever
  et~al\mbox{.}}{2014}]%
        {conf/nips/SutskeverVL14}
\bibfield{author}{\bibinfo{person}{Ilya Sutskever}, \bibinfo{person}{Oriol
  Vinyals}, {and} \bibinfo{person}{Quoc~V. Le}.}
  \bibinfo{year}{2014}\natexlab{}.
\newblock \showarticletitle{Sequence to Sequence Learning with Neural
  Networks.}
\newblock \bibinfo{journal}{\emph{NIPS}} (\bibinfo{year}{2014}),
  \bibinfo{pages}{3104--3112}.
\newblock


\bibitem[\protect\citeauthoryear{Toshev and Szegedy}{Toshev and
  Szegedy}{2014}]%
        {Toshev_2014_CVPR}
\bibfield{author}{\bibinfo{person}{Alexander Toshev} {and}
  \bibinfo{person}{Christian Szegedy}.} \bibinfo{year}{2014}\natexlab{}.
\newblock \showarticletitle{DeepPose: Human Pose Estimation via Deep Neural
  Networks}. In \bibinfo{booktitle}{\emph{CVPR}}. \bibinfo{pages}{1653--1660}.
\newblock


\bibitem[\protect\citeauthoryear{Vinyals and Le}{Vinyals and Le}{2015}]%
        {journals/corr/VinyalsL15}
\bibfield{author}{\bibinfo{person}{Oriol Vinyals} {and}
  \bibinfo{person}{Quoc~V. Le}.} \bibinfo{year}{2015}\natexlab{}.
\newblock \showarticletitle{A Neural Conversational Model.}
\newblock \bibinfo{journal}{\emph{CoRR}}  \bibinfo{volume}{abs/1506.05869}
  (\bibinfo{year}{2015}).
\newblock
\urldef\tempurl%
\url{http://dblp.uni-trier.de/db/journals/corr/corr1506.html\#VinyalsL15}
\showURL{%
\tempurl}


\bibitem[\protect\citeauthoryear{Yi~Li}{Yi~Li}{2017}]%
        {fcis2017li}
\bibfield{author}{\bibinfo{person}{Jifeng Dai Xiangyang Ji Yichen~Wei Yi~Li,
  Haozhi~Qi}.} \bibinfo{year}{2017}\natexlab{}.
\newblock \showarticletitle{Fully Convolutional Instance-aware Semantic
  Segmentation}.
\newblock \bibinfo{journal}{\emph{CVPR}} (\bibinfo{year}{2017}).
\newblock


\bibitem[\protect\citeauthoryear{You, Zhang, Hsieh, Demmel, and Keutzer}{You
  et~al\mbox{.}}{2017}]%
        {You2017}
\bibfield{author}{\bibinfo{person}{Yang You}, \bibinfo{person}{Zhao Zhang},
  \bibinfo{person}{Cho{-}Jui Hsieh}, \bibinfo{person}{James Demmel}, {and}
  \bibinfo{person}{Kurt Keutzer}.} \bibinfo{year}{2017}\natexlab{}.
\newblock \showarticletitle{{ImageNet} Training in Minutes}.
\newblock \bibinfo{journal}{\emph{CoRR}}  \bibinfo{volume}{abs/1709.05011}
  (\bibinfo{year}{2017}).
\newblock


\bibitem[\protect\citeauthoryear{Yu, Eversole, Seltzer, Yao, Kuchaiev, Zhang,
  Seide, Huang, Guenter, Wang, Droppo, Zweig, Rossbach, Gao, Stolcke, Currey,
  Slaney, Chen, Agarwal, Basoglu, Padmilac, Kamenev, Ivanov, Cypher,
  Parthasarathi, Mitra, Peng, and Huang}{Yu et~al\mbox{.}}{2014}]%
        {cntk}
\bibfield{author}{\bibinfo{person}{Dong Yu}, \bibinfo{person}{Adam Eversole},
  \bibinfo{person}{Mike Seltzer}, \bibinfo{person}{Kaisheng Yao},
  \bibinfo{person}{Oleksii Kuchaiev}, \bibinfo{person}{Yu Zhang},
  \bibinfo{person}{Frank Seide}, \bibinfo{person}{Zhiheng Huang},
  \bibinfo{person}{Brian Guenter}, \bibinfo{person}{Huaming Wang},
  \bibinfo{person}{Jasha Droppo}, \bibinfo{person}{Geoffrey Zweig},
  \bibinfo{person}{Chris Rossbach}, \bibinfo{person}{Jie Gao},
  \bibinfo{person}{Andreas Stolcke}, \bibinfo{person}{Jon Currey},
  \bibinfo{person}{Malcolm Slaney}, \bibinfo{person}{Guoguo Chen},
  \bibinfo{person}{Amit Agarwal}, \bibinfo{person}{Chris Basoglu},
  \bibinfo{person}{Marko Padmilac}, \bibinfo{person}{Alexey Kamenev},
  \bibinfo{person}{Vladimir Ivanov}, \bibinfo{person}{Scott Cypher},
  \bibinfo{person}{Hari Parthasarathi}, \bibinfo{person}{Bhaskar Mitra},
  \bibinfo{person}{Baolin Peng}, {and} \bibinfo{person}{Xuedong Huang}.}
  \bibinfo{year}{2014}\natexlab{}.
\newblock \bibinfo{booktitle}{\emph{An Introduction to Computational Networks
  and the Computational Network Toolkit}}.
\newblock \bibinfo{type}{{T}echnical {R}eport}.
\newblock


\end{thebibliography}

\newpage

%
\appendix
\section{Details of Experimental Setups}

We herein provides the detailed setups for experiments whose results are provided in the main text.

\subsection{Distributed Training}

In the experimental evaluation at Section~\ref{sec:evalmn}, we used development branches based on Chainer 3.0.0rc1 and ChainerMN 1.0.0 \footnote{At the time when we ran this evaluation ChainerMN was released as a plugin library to Chainer, while recently it has been merged to the mainline of Chainer.}.
As the underlying communication libraries, we used NCCL 2.0.5 and OpenMPI 1.10.2.

We used an in-house cluster that consists of 128 nodes. Each node is equipped with two Intel Xeon E5-2667 processors (3.20 GHz, eight cores), 256-GB memory, and eight NVIDIA Tesla P100 GPUs. All nodes are interconnected by the Mellanox Infiniband FDR.

The per-worker minibatch size was 32 and the total minibatch size was 32k with 1024 workers.
Computations were generally performed in single precision;
 to reduce the payload size, we used half-precision floats for communication.

\end{document}